\documentclass[a4paper]{article}
\usepackage[utf8]{inputenc} 
\usepackage[T1]{fontenc}    
\usepackage{hyperref}       
\usepackage{url}            
\usepackage{booktabs}       
\usepackage{amsfonts}       
\usepackage{amsmath}
\usepackage{nicefrac}       
\usepackage{microtype}      
\usepackage{multirow}
\usepackage{svg}
\usepackage{subfig} 
\usepackage[export]{adjustbox}
\usepackage{fullpage}

\begin{document}

\title{Alphazzle: Jigsaw Puzzle Solver with Deep Monte-Carlo Tree Search}
\date{September 2020}

\author{Marie-Morgane Paumard$^1$, Hedi Tabia$^2$, David Picard$^1$ \\ \small{$^1$ LIGM, Ecole des Ponts, CNRS, Univ Gustave Eiffel, Marne-la-vallée, France} \\ \small{$^2$ IBISC, Univ. Evry, Université Paris-Saclay, France}}
%


\maketitle

\begin{abstract}
    Solving jigsaw puzzles requires to grasp the visual features of a sequence of patches and to explore efficiently a solution space that grows exponentially with the sequence length. Therefore, visual deep reinforcement learning (DRL) should answer this problem more efficiently than optimization solvers coupled with neural networks. Based on this assumption, we introduce Alphazzle, a reassembly algorithm based on single-player Monte Carlo Tree Search (MCTS). A major difference with DRL algorithms lies in the unavailability of game reward for MCTS, and we show how to estimate it from the visual input with neural networks. This constraint is induced by the puzzle-solving task and dramatically adds to the task complexity (and interest!). We perform an in-deep ablation study that shows the importance of MCTS and the neural networks working together. We achieve excellent results and get exciting insights into the combination of DRL and visual feature learning.
\end{abstract}


\section{Introduction}

\begin{figure}
  \centering
  \includegraphics[width=1\columnwidth]{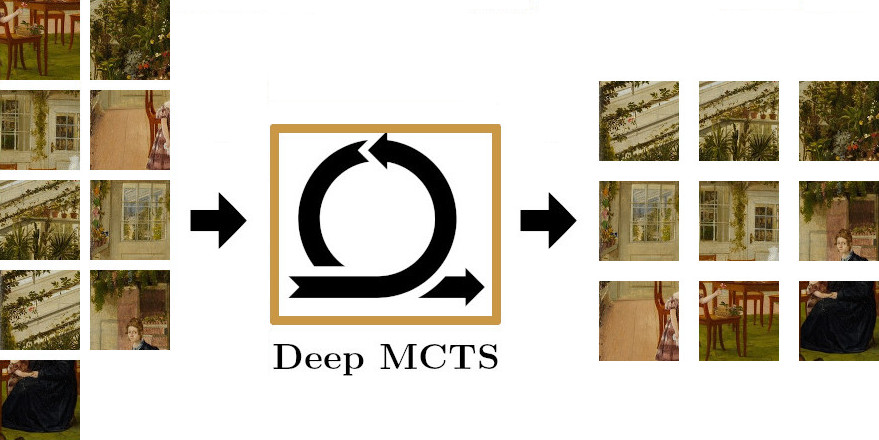}
  \caption{Example of a jigsaw puzzle task.}
  \label{fig:intro}
\end{figure}

The standard jigsaw puzzle-solving task consists in predicting the position of nine square patches to form a $3 \times 3$ image (Figure~\ref{fig:intro}). For the computer vision community, it is a very interesting task. First, as a pretext task, it improves neural networks performances \cite{doersch2015unsupervised, noroozi2018boosting, carlucci2019domain, kim2018learning}. Second, as a standalone goal, it is one of the few tasks that combine two traditionally opposed aspects of artificial intelligence: visual understanding and choice space exploration. Robotics exhibits a similar positioning between these two domains, with the differences that the choice space is much larger and that there are many optimal action path. In contrast, puzzle-solving occurs in a controlled environment characterized by a narrow choice space and a unique correct path---given an ordered set of patches. Thus, it exhibits excellent properties to study in depth this combination of visual understanding with choice space exploration. It also has a wide range of applications, from few-shot learning to genome biology, forensic, and archaeology. Deep reinforcement learning algorithms are well suited to such tasks and are able to defeat brilliantly human players on complex board games. Therefore, we make the hypothesis that deep reinforcement learning is more suited to solve jigsaw puzzles than deep learning coupled with a solver \cite{noroozi2016unsupervised, paumard2020deepzzle, bridger2020solving}.

In this article, we propose Alphazzle, a deep Monte Carlo Tree Search (MCTS) algorithm that manages images as input and solves jigsaw puzzles iteratively. Alphazzle is derived from AlphaZero and display the following characteristics: Its MCTS is adapted to single-player games. It relies on neural networks to interpret the images and estimate the reward. Contrarily to board games, this reward cannot be deduced from the rules while playing. This last constraint is inspired by some real jigsaw puzzle tasks, such as archaeological patches reconstruction, where we cannot formulate explicit rules to determine if a puzzle is well solved. Therefore, in addition to perform the reassembly, the algorithm should also correctly assess if an image is correctly reassembled.
Our setup provides a controlled environment for jointly learning visual representations and action space exploration on which Alphazzle sets a clean baseline that the community car compare to.

This paper is organized as follows: In Section~\ref{sec:soa}, we review the literature on deep learning-based puzzles solvers and deep reinforcement learning with MCTS. Then, we detail our method in Section~\ref{sec:mth} and present our results in Section~\ref{sec:exp}.


\section{Related work}
\label{sec:soa}

In this section, we present the jigsaw puzzle solvers based on deep learning. Next, we introduce the deep reinforcement learning algorithms based on Monte Carlo Tree Search, for one and two-player games.

\subsection{Solving puzzles with deep learning}
\label{sec:soa:dl}

Traditional computer vision algorithms based solely on pattern recognition such as~\cite{andalo2016psqp, chen2018new, gur2017square, hammoudeh2017clustering, sivapriya2018automatic} manage hundreds of pieces but do not generalize. For example, they show low resiliency to erosion and cannot make relation-based hypotheses such as placing the sky above the ground.

In 2015, Doersch et al.~\cite{doersch2015unsupervised} introduced a pretext task for classification and detection: they used a jigsaw puzzle solver to endow a neural network with a sense of spatial semantics. Their network solves $3 \times 3$ puzzles by predicting the relative positioning of a lateral patch compared to a given central patch, among $8$ positions. Their experiment does not include the reassembly of the patches.

Henceforth, two types of reassembly methods prevail. On the one hand, there are those inspired by Doersch et al., such as Ostertag and Beurton-Aimar~\cite{ostertag2020matching}, Paumard et al.~\cite{paumard2020deepzzle}, and Bridger et al.~\cite{bridger2020solving}. Their neural network predicts the relative position of each couple of patches. Then, they use a graph to minimize joint placement prediction. The first two compare each patch to the central one, so the reassembly is not very precise; Bridger et al. significantly increased both state-of-art scores on big puzzles and computation time. On the other hand, Carlucci et al.~\cite{carlucci2019domain} and Favaro~\cite{noroozi2016unsupervised}, do not compare the lateral patches to the central one. They solve puzzles by finding the correct permutation to perform on the patches, were each permutation is a class of a classification problem. Therefore, the number of classes grows exponentially, so authors usually drastically restrict the available permutations to a few hundreds.

Last, Wei et al. \cite{wei2019iterative} introduced iterative reassemblies with deep learning. They combine losses to take into account the pairwise patches comparison and the global reassembly score. Then, the permutation is applied, and the new reassembly is re-evaluated, iteratively until a stop criterion is reached. This method takes the best of the previous algorithms as it can cover all permutations and solve the puzzle based on all the relations between the patches. However, it is still limited by computation power.

Overall, the computer vision literature on solving jigsaw puzzle is either interested in the pretext task or in producing correct reassembly. 
In contrast, we are interested in evaluating how to learn visual features for action space exploration, for which we think solving visual puzzle is the perfect controlled environment.

\subsection{Monte Carlo Tree Search}

Monte Carlo Tree Search is a two-agent heuristic search algorithm in a tree, introduced by \cite{coulom2006efficient} for Go game \cite{gelly2006exploration} and proven to be guaranteed to converge \cite{kocsis2006bandit}. Shortly after, algorithms using both MCTS and reinforcement learning emerged \cite{silver2009reinforcement, browne2012survey, vodopivec2017monte}. More recently, is has been combined with deep reinforcement learning, leading to AlphaGo \cite{silver2016mastering}, AlphaGo Zero \cite{silver2017masteringAGZ}, AlphaZero \cite{silver2017mastering}, ExIt \cite{anthony2017thinking} and MuZero \cite{schrittwieser2019mastering} algorithms.

MCTS should be adjusted to address single-player games; we count two main issues. First, most single-player games reward are not bounded, while MCTS is adapted for rewards among $\{-1,0,1\}$. Second, the values found during the tree search are a lower bound of the optimal value, so the selection step must be adjusted.
Schadd et al. \cite{schadd2012single} introduced SP-MCTS, which improves among other things the performances of the selection phase. They also create a tree per move rather than a tree per game and change MCTS selection formula to compensate for the non-adverse aspect.
Baier and Winands \cite{baier2012nested} propose a recursive MCTS that solves single-players games such as Bubble Breaker. It is based on NMCS \cite{cazenave2009nested}, which also inspired Rosin's NRPA \cite{rosin2011nested} that was able to solve Morpion Solitaire and construct crossword puzzles.
Orseau et al. \cite{orseau2018single} introduce two tree-search algorithms for single-player games. The first one is adapted for “needle-in-a-haystack” problems, i.e., problems for which the number of correct solutions is very limited. It derivates from Levin’s search \cite{levin1973universal}. The second one is well-suited for problems where many paths lead to a goal.
Seify and Buro \cite{seify2020single} proposes a variant of MCTS which is adapted for games with unbounded rewards. Like Schadd et al. \cite{schadd2012single}, they propose a variant of the MCTS selection formula.

Last, we present some single-player games solvers that use deep reinforcement learning and tree search. Arfaee et al. \cite{arfaee2011learning} proposed in 2011 to combine a neural network with the tree search algorithm IDA*. Laterre et al. \cite{laterre2018ranked} proposed R2, an algorithm for single-player games. It addresses the issue of unbounded reward with a relative performance metric. McAleer et al. \cite{mcaleer2018solving} proposed DeepCube, a solver for the Rubik’s cube. They pre-train a two-headed neural network. After the network is trained, it is combined with MCTS to effectively solve the Rubik’s cube. Agostinelli et al. \cite{agostinelli2019solving} continued the work of McAleer et al. and proposed DeepCubeA. In brief, they replaced MCTS with a weighted A* search. Their algorithm solves Rubik’s cube and 8-puzzle (also known as gem puzzle and mystic square).

As we can see, in the MCTS related literature, most research has focus on problems where the relevant information is readily available in semantic form (like the board state of a game of Go).
In contrast, we are interested in exploring action spaces where the relevant information has to be extracted from the input signal. 
We argue that solving jigsaw puzzle is an excellent task for that since semantically sound visual features have to be learned.
In comparison to video games, puzzle solving also provide a much more controlled environment where he action space is much more limited, allowing to better analyze the results.


\section{Methods}
\label{sec:mth}

In this article, we present Alphazzle, an iterative reassembly method for jigsaw puzzles. Rather than opting for a recursive model such as RNN \cite{rumelhart1987rnn}, we cast this task as a planning problem, and thus apply a model-based reinforcement learning framework inspired by AlphaZero \cite{silver2017mastering}. 

\subsection{Overview}
\label{sec:az:overview}

In this subsection, we focus on the interaction between MCTS and the neural network. We start by presenting two-player games with deep reinforcement learning, which allows us to detail the notations, and AlphaZero, which gives an overview of the interactions between the tree search and the deep learning. Then, we introduce the rules of the jigsaw puzzle game.


\subsubsection{Simplified framework for two-player games}

Two agents $(g_1, g_2)$ play a turn-based game. The game is defined by a set of hard-coded rules, which includes the initial board state $s_0$, the available actions given a current state $s_t$, the end game criteria, and the scoring function $r(s_{t\_max})\in\{-1,0,1\}$ (i.e., the reward), that is computed at the end of the game $t\_max$. If the score is null, there is a tie; if $r(s_{t\_max})=+1$, agent $g_1$ won the game, and vice versa. Note that $t\_max$ varies between two games.

At each turn $t$, one of the agents chooses the available action $a_t$ that is presumed to minimize the opponent’s final gain: $g_1$ aims to maximize $r(s_{t\_max})$ and $g_2$ wants to minimize it.

The next action choice is based on policy $\pi_\theta(a_t|s_t)$, where $\theta$ are the parameters of the neural network $P$ that returns the policy. As some games’ duration can be very long, getting samples of games and associated rewards is not feasible in a reasonable amount of time. Therefore, having an estimator of the value function is a must. The value function is the sum of expected rewards values, given a current state $s_t$: $v(s_t)=\mathbb{E}(r(s_{t\_max})|s_{t})$. Consequently, there is a neural network $V$ in charge of learning $v(s_t)$, and guiding the optimization of $\pi_\theta$.


\subsubsection{AlphaZero algorithm}

To the framework presented above, AlphaZero adds a planning algorithm, Monte Carlo Tree Search (MCTS), which explores many compelling actions’ further effects. Instead of selecting the action $a_t$ that maximizes $v(s_t)$ according to $V$ and $P$, the agent performs simulations of what could happen according to its action. Therefore, it selects its action based on the value of the most promising explored state.

MCTS returns the policy $\pi_{MCTS}(a_t|s_t)$. During inference, the $g_1$ selects the best action and update the state; then, $g_2$ applies MCTS from the state $s_{t+1}$. During the learning phase, an exploration trade off is provided to the agents, and the neural networks are engaged in reinforcement learning, playing until the accuracy is acceptable.

\begin{figure}
    \centering
    \includegraphics[width=0.7\linewidth]{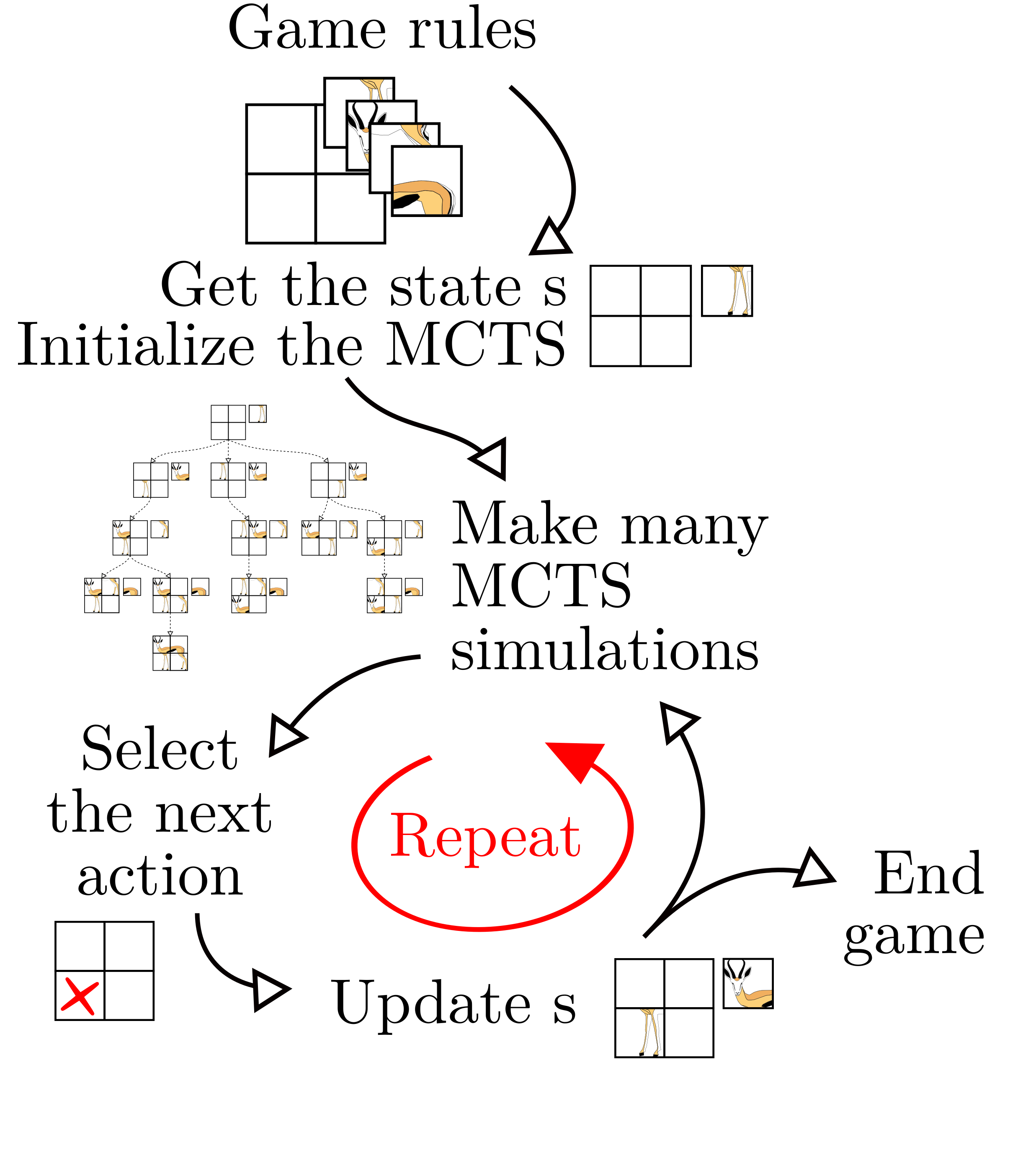}
    \caption[Outline of Alphazzle]{Alphazzle outline.}
    \label{fig:az:overview}
\end{figure}
Our core algorithm (Figure \ref{fig:az:overview}) reproduces AlphaZero. 


\subsubsection{Jigsaw puzzle rules and formalization}

A single agent plays the game.

\paragraph{States and actions} The current (board) state $s_t$ is described by the arbitrarily ordered set of patches to place and the current partial reassembly, obtained from the already placed patches. The observable state only contains the next patch to place $x_{f,t}$ and the partial reassembly $x_{r,t}$. 

The available actions $A_t$ are those that assign the next patch to any empty position. When the agent performs action $a_t\in A_t$, the patch is added to the partial reassembly and the state is updated to $s_{t+1}$. The environment is deterministic, so we know what $s_{t+1}$ is from $s_t$ and $a_t$.

It is equivalent to having an unordered set of patches in $s_t$ and letting MCTS selects which patches to place where. The width of the tree increases strongly, but the ratio of correct paths is unchanged. We choose to fix the patches’ order to reduce the tree size. The downside is that, as some orders as easier to solve than others, finding a correct reassembly may be more difficult with our setup. To compensate for this, we solve several times the same puzzle with different patches’ orders.

\paragraph{Initialization and endgame} At first, $x_{r,t=0}$ is the zero matrix. Its size is fixed and depend on the number of patches, on their size, and on the gap between patches size. To differentiate black patches from empty locations, we also initialize to $-1$ a dictionary $d_t$ which matches the positions $j \in [0\mathrel{{.}\,{.}}\nobreak p]$ in the reassembly to the indexes of the patches $i \in [0\mathrel{{.}\,{.}}\nobreak f]$, , where $f$ is the number of lateral patches and $p$ the number of position.

The game ends when all the positions are filled up:
$\forall j \in [0\mathrel{{.}\,{.}}\nobreak p] , d_{t_{max}}[j]\neq -1$,
or as soon as all the patches are placed:
$x_{f,t_{max}}=None$.
Consequently, the depth of the tree spanning the action space is bounded by $\min(p,f)$, which is not the case for AlphaZero’s games, as games of very different length can be generated. 

\paragraph{Reward} \label{sec:az:reward} For the reward, we can use three metrics to evaluate how correct the game is: the percentage of correct neighbors, the percentage of well-placed patches and whether the puzzle is solved. All of them are bounded by one, so we do not face unbounded reward described by \cite{schadd2012single}.

For the training, we opt for the binary solved-puzzle reward: we expect this reward encourages MCTS to focus on the solution and discarding all wrong reassemblies, even those with a high percentage of correct neighbors. The downside is the wrong reassemblies are all considered equivalent, i.e., they are not ordered by their number of mistakes.

For comparison, we endow Alphazzle’s MCTS with two reward modes.
The first one is based on ground-truth: if the final reassembly dictionary equals the solution dictionary, then $r(s_{t_{max}})=1$, else $r(s_{t_{max}})=0$.
The second one is based on an automatic assessment of the realism of the reassembly by a neural network. In the real world, jigsaw puzzles from archaeology or medicine do not come with their solution, so experts must evaluate if the reassembly is correct.
Compared to AlphaZero and other deep reinforcement learning algorithms, this is new. We describe this reward mode as “the (ground-truth) reward is not available to MCTS.”. We use an evaluator who has learned how to classify the correct reassembly, which is $V$.


\subsection{Monte Carlo Tree Search}
\label{sec:az:mcts}

MCTS visits $N_{visits}$ nodes from the current state $s_t$, and returns $a_t$, the most promising action for the step $t$. The current player applies it. Then, MCTS explores $N_{visits}$ nodes, starting from $s_t+1$ and returns $a_{t+1}$, and so on until the end of the game. After $N_{visits}$ nodes visited, MCTS returns the policy $\pi_{MCTS}(a_t|s_t)$. MCTS applies the four following steps $N_{visits}$ times: Selection, Expansion, Simulation and Backpropagation.

\begin{figure}
    \centering
    \includegraphics[width=0.9\linewidth]{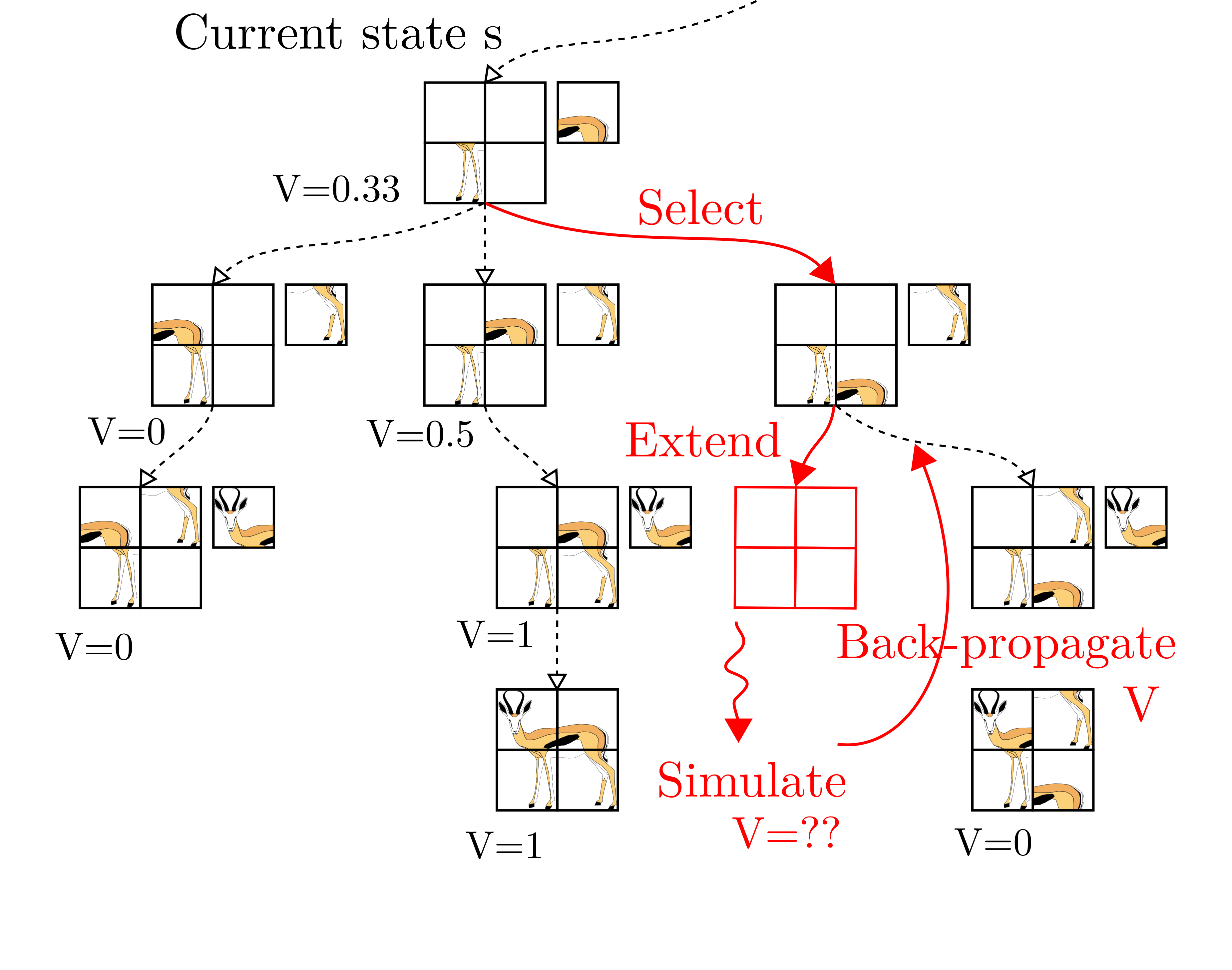}
    \caption[Outline of MCTS, applied to puzzles]{Example of MCTS simulations applied to puzzles. In this example, the states are clearly shown, as well as the state of the value function, which is the expectation to find the correct reassembly from the current node.}
    \label{fig:soaaz:ourmcts}
\end{figure}

Figure \ref{fig:soaaz:ourmcts} shows our MCTS applied to a 2×2 jigsaw puzzle. We display the value of $v(s_t)$, the expectation of finding the solution from each state $s_t$.
The possible states are represented with a tree, whose branches are the actions. At the beginning of MCTS, there is only one node, $s_t$. Each visit add nodes to the tree. Each row of the tree represents a turn consisting in placing a patch.

\subsubsection{Selection}

The Selection phase enables picking a node among the leaves according to a vector $U(a|s_t)$, named from Upper Confidence Bound (UCB \cite{auer2002finite}. It assigns values to each available action from the state $s_t$. Then, the best action according to $U(a|s_t)$ is applied recursively until a leaf state is reached. If the leaf has children already, the Selection algorithm can either stop or selects a child.

In 2006, Kocsis and Szepesvári proposed Upper Confidence bounds applied to Trees (UCT) \cite{kocsis2006bandit}, a selection strategy derived from UCB. It states:
\begin{equation}
    \label{eq:az:01}
    \forall a \in A_t, U(a|s_t) = Q(a|s_t)+C\cdot\sqrt{\frac{\log N(s_t)}{N(a|s_t)}},
\end{equation}
where $A_t$ is the set of available actions at step $t$,
$Q(a|s_t)$ is the expected value of the available actions (see Backpropagation),
$C$ is the exploration trade-off constant,
$N(s_t)$ is the number of visits to the node associated with $s_t$,
$N(a|s_t)$ is the number of times the available actions have been taken from the state $s_t$. Note that $\forall a \in A_t, N(a|s_t) = N(s_{t+1|a})$.

This strategy $U$ has been adapted to single-player games by Schadd et al. \cite{schadd2012single}:
\begin{equation*}
    \forall a \in A_t, U_{SP}(a|s_t) = U(a|s_t)+W\cdot Q_{max}(a|s_t)+\sigma(a_t|s_t).
\end{equation*}
They made two modifications on \ref{eq:az:01}:
\begin{itemize}
    \item $W\cdot Q_{max}(a|s_t)$ is a fraction of the maximum value obtainable from the action $a$ applied from $s_t$. This term indicates that it is relevant to focus not only on $Q(a|s_t)$, the average value that can be obtained from $a$, but also on the maximum value that can be derived from it. This is possible because there are no opponents: in two-player games, if the first player chooses the action that has a bad average but a maximum value, the second player will steer the game so that the maximum value is not reached. Schadd et al. set $W=0.02$. Jacobsen et al. \cite{jacobsen2014monte} recommend using $(1-\lambda)\cdot Q(a|s_t)+\lambda\cdot Q_{max}(a|s_t)$ rather than $Q(a|s_t)+W\cdot Q_{max}(a|s_t)$.
    \item $\sigma(a_t|s_t)$ is the standard deviation estimate. This term is pertinent in case of game with unbounded rewards.
\end{itemize}
AlphaGo \cite{silver2016mastering} introduced PUCT, derived from PUCB \cite{rosin2011multi}:
\begin{equation}
    \label{eq:az:02}
    U(a|s_t) = Q(a|s_t)+C\cdot \pi_\theta(a|s_t) \cdot \frac{\sqrt{ N(s_t)}}{1+N(a|s_t)},
\end{equation}
where $\pi_\theta(a|s_t)$ is the policy returned by the neural network $P$ that predicts the actions.

In Alphazzle, we mostly use Equation \ref{eq:az:02}.


\subsubsection{Expansion}

The Expansion phase occurs after the Selection and enables appending one or several nodes to the tree. An example of an Expansion strategy is to select an unexplored node randomly.
In AlphaZero, one node is expanded at each iteration. The expanded node is obtained from the best $U(a|s_t)$. Indeed, in PUCT, the predictors allows $U(a|s_t)$ to select an unexplored node. Therefore it is used for both the Selection and Expansion phases.

In Alphazzle, we use AlphaZero Expansion.


\subsubsection{Simulation}

The Simulation phase aims to find a possible value obtained from the expanded node. The objective is to make (more) accurate this node’s value and its predecessors during the Backpropagation phase. An example of a Simulation strategy is to select actions randomly until an endgame is reached. In most cases, however, a handmade policy guides the Simulation. For example, Schadd et al. \cite{schadd2012single} propose two policies for SameGame, which promotes the creation of large groups of color.

In AlphaZero, Silver et al. replace the Simulation phase with a neural network $V$, which predicts the expected value from the expanded node. When their MCTS reach an endgame node, the ground-truth reward is returned.

In Alphazzle, we use AlphaZero Simulation, except that the endgame ground-truth reward may be replaced by the predicted reward, depending on our experimental settings. 


\subsubsection{Backpropagation}
\label{sec:az:backprop}

The Backpropagation phase updates $Q(a|s_t)$, $N(a|s_t)$ and $N(s_t)$ of each node visited.

Given $v(s_t)$ the value of the leaf node, we initialize $Q(a|s_t) = v(s_t)$, $N(a|s_t) = 1$ at the first visit of the node. At each next visit, we perform the update:
\begin{equation*}
    Q(a|s_t) \gets \frac{N(a|s_t)\cdot Q(a|s_t) +v(s_t)}{N(a|s_t)+1},
\end{equation*}
\begin{equation*}
    N(a|s_t) \gets N(a|s_t)+1, \text{ and } N(s_t) \gets N(s_t)+1.
\end{equation*}


\subsection{Deep Reinforcement Learning}
\label{sec:az:drl}

Like McAleer et al. suggested in \cite{mcaleer2018solving}, we pre-train our neural networks on handcrafted tasks. We expect such pre-training enables them to be more accurate. After the pre-training, we integrate them into our MCTS algorithm. Then, we fine-tune on training examples that the MCTS was not able to solve.

Another difference with AlphaZero is that our $P$ and $V$ use visual inputs and thus have to extract meaningful information from these noisy signals. In contrast, AlphaZero directly uses the board state, which is a clean semantic input. For that reason, we use two image datasets: one for training $P$ and $V$ and one for evaluating our algorithm.

\subsubsection{Pre-training $P$}
The neural networks $P$ predicts the best actions from a patch’s image and a partial reassembly image. We generate inputs from our puzzle dataset: we make correct partial reassemblies and select patches to place among the remaining patches. Note that $P$ is not trained on wrong partial reassemblies, because there may be no correct answer for the patch to place (i.e., its position is already taken).

The neural network $P$ is trained using categorical cross-entropy to predict the patch position, outputting an estimate $\vec{p}_s$ of the policy.

\subsubsection{Pre-training $V$}
The neural networks $V$ predicts the expected reward value that can be reached from the state $s_t$. Therefore, its input is a reassembly, either partial or complete. When the reassembly is complete, or when only one patch is remaining, $V$ predicts the reward: 0 if there is at least a mistake and 1 otherwise. When the reassembly is partial, $V$ is trained to predict 0 if there is at least a mistake. If the partial reassembly is correct, it can lead to wrong and correct reassemblies; therefore we train $V$ to predicts:
\begin{equation}
    \label{eq:az:03}
    v(s_t)=0.5+0.5\cdot\frac{i}{f-1},
\end{equation}
where $f$ is the number of patches, and $i$ is the number of well-placed patches in state $s_t$. We choose this value because it indicates the confidence in the prediction: an empty puzzle predicted value is the $0.5$, because we have no information.

\subsubsection{Architectures}
In our experiments, we use a WideResNet (WRN) \cite{zagoruyko2016wide} initialized with random weights for the features extractor part of $P$. Each input goes through the same feature extractor, thanks to shared weights. Then, each goes to a different multi-layer perceptron made of 6 successive fully-connected layers of size 512, alternating with ReLU functions. The two outputs are concatenated and given to a shallower perceptron that predicts the patch position (2 fully-connected layers, a ReLU between them, and a softmax at the end). The classes correspond to the puzzle positions, and $P$ is trained by sampling random partial reassemblies.

The neural network $V$ is trained using MSE loss.
Its architecture borrows the WRN (or ResNet) from $P$, followed by an MLP.

\subsubsection{MCTS-based fine-tuning}
Finally, we introduce the fine-tuning of $P$ and $V$ that could occur while solving puzzles. 

This mining of training examples is akin to active learning, where the learning focuses on more important examples. Indeed, during the pre-training, reassemblies are sampled uniformly, while it is not the case of the nodes explored by MCTS. Thus, we suggest to fine-tune the networks on the nodes that are likely to be visited. This process differs from AlphaZero: in a two-player game, there is always a set of actions that led to victory and thus can be used to reinforce the $PV$ network, especially because $V$ has access to the ground truth at terminal nodes.

We use the states obtained from all the choices made after the policy $\pi_{MCTS}(a_t|s_t)$. In AlphaZero, the neural networks learn to reproduce actions that lead to the victory of an agent. In our case, we cannot deploy such learning from the opponent; therefore, we learn the ground-truth for $P$ (even if the position is already occupied) and the value from Equation \ref{eq:az:03} for $V$. As such, we can only learn on the same supervised training set that is used during pre-training.


\section{Results}
\label{sec:exp}

In this section, we first introduce our experiments. Then, we detail our neural networks pre-training performances and the MCTS performances. Last, we present our results on the jigsaw puzzle task. We chose to focus on the conclusion of our numerous experiments in the main text, while extensive results are available in appendix.


\subsection{Experiments}

We program the neural networks with PyTorch library. Table \ref{tab:exp} shows the standard parameters for our experiments.

\begin{table}
\begin{center}
\begin{tabular}{|l|c|}
\hline
    Feature extraction & WRN \\
    Optimizer & Adam \\
    Learning rate & 0.001 \\
\hline\hline
    Patch size & 40 \\
    Patch per size & 3 \\
    Space size & 4\\
\hline\hline
    Selection & PUCT \\
    Reward & Predicted \\
    Action choice & $N(a|s_t)$ \\
\hline
\end{tabular}
\end{center}
\caption{Summary of the experiments parameters.}
\label{tab:exp}
\end{table}

\paragraph{Dataset} We train our neural network on MET (10,000 training images and 2,000 validation images) \cite{paumard2018image}. At each epoch during training, we use different crops within the images. For each puzzle, we online train on a single pair of fragment, partial reassembly. We normalize the values between -1 and 1.


\subsection{Pre-training performances}
\label{sec:azr:pretrain}

\begin{table}
\begin{center}
\begin{tabular}{|l|c|c|}
    \hline
    & P & V \\
    \hline\hline
    Validation accuracy & 69.91 \%  & 88.46 \% \\
    \hline
\end{tabular}
\end{center}
  \caption{Validation accuracy after 100 epochs.}
  \label{tab:azr:resnet}
\end{table}

\begin{figure}
    \centering
    \subfloat{\includegraphics[width=0.45\linewidth]{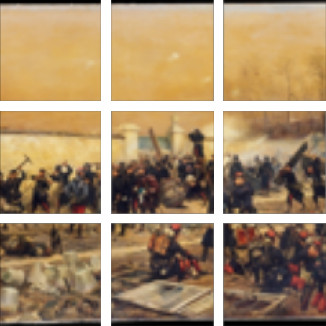}}\hfill
    \subfloat{\includegraphics[width=0.45\linewidth]{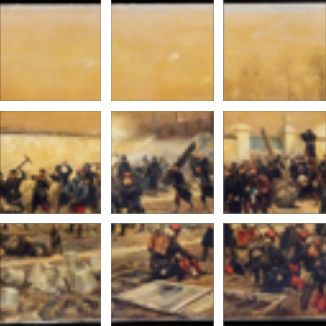}}
    \caption{Comparison of a correct reassembly (left) and a incorrect one with one inversion (right).}
    \label{fig:azr:inversion}
\end{figure}

Table \ref{tab:azr:resnet} shows the validation accuracy of $P$ and $V$ after pretraining. Besides, we ran several settings comparison to understand the impact of the number of patches per side, the space size, the patch size, the already placed patches and detail our results in appendix. Among other analysis, we study the accuracy of $V$ on complete puzzles. On average, on our pre-training distribution, we obtain 96.77\%. Half of this distribution are perfect reassemblies (i.e., $9$ well-placed patches), the other half are reassemblies with any number of mistakes (i.e., $<7$ well-placed patches). Interestingly, it is difficult for $V$ to evaluate puzzles with only one inversion, but as soon as three patches are inverted, the precision goes back up. Figure \ref{fig:azr:inversion} shows an example of a two-patches inversion and illustrate the difficulty of the task: the right image may seem correct. Moreover, many images from MET have a plain background and interchangeable patches, but our metric does not take it into account.


\subsection{MCTS performances}
\label{sec:azr:mcts}

After running a grid-search, we found that MCTS is not too much sensitive on C and V and we set $C=1$ and $N_{visits}=10^3$, as exploring more nodes would take more than one minute per puzzle to solve. The full parameter study is available in appendix.

\subsubsection{Predicted reward versus ground-truth reward}

\begin{table}
\begin{center}
  \begin{tabular}{|l|c|c|c|}
    \hline
    Reward & Patch & Neighbor & Puzzle\\
    \hline\hline
    Ground-truth & 78.14 \% & 79.92 \% & 70.55 \% \\
    Predicted & 55.63 \% & 58.93 \% & 14.55 \% \\
    \hline
\end{tabular}
\end{center}
  \caption{Reassemblies scores: endgame rewards comparison.}
  \label{tab:azr:endgame}
\end{table}

Table \ref{tab:azr:endgame} shows that when using the predicted reward, scores drop by 20\% for patch-wise and neighbor-wise metrics, and by 55\% for the reassemblies, which is a significant difference. We assume that this is due to the accuracy issue of $V$ on complete puzzles. On the one hand, with ground-truth reward, if MCTS finds a path with $r(a,s)=1$, it is the correct reassembly, and so MCTS will select it. On the other hand, with predicted reward, MCTS may finds paths with very high values of $r(a,s)$, but that leads to wrong reassemblies.

\subsubsection{Deactivation of $P$ and $V$}
To measure the importance of $P$ and $V$ for MCTS, we compare the impact of deactivating them, i.e., replacing $P$ by a unit vector or $V$ by a constant. 
The sappendix presents the full results. It appears that MCTS can cope with the absence of either $P$ or $V$, but not the lack of both. We note that deactivating $P$ has more impact than replacing $V$ by 1 during the game. If we keep $P$ and $V$ but stop predicting the reward from the middle-game, the results are close to the baseline. However, if we remove $P$ or $V$ and the middle-game predictions, the results drop. Especially if we remove $P$, we are not able to reassembly any puzzle. On the contrary, if we only deactivate $V$ for the endgame reward and predict 1, the neural network still picks pertinent reassemblies, although it considers them all equivalent.

\subsubsection{Greedy neural networks}
We compare our results with a baseline that uses a greedy exploitation by taking the argmax of $P$ or $V$ at each step, without MCTS. We show the results in Table \ref{tab:azr:greedy}. When $P$ solves a puzzle by itself, it is shown couples of patches and partial reassemblies, starting from the empty reassembly. Then, the patches are placed according to $P$’s predictions, updating the reassembly. Without MCTS, the results’ quality drops, which shows the importance of exploring alternative reassemblies with MCTS. Similarly, $V$ evaluates all the partial reassemblies that can be obtained from the current state. Then, it selects the action leading to the best reassembly according to its predictions. It is interesting to see that $V$ is a better action predictor than $P$ alone.

\begin{table}
\begin{center}
  \begin{tabular}{|l|c|c|}
    \hline
    Reward & Patch-wise & Puzzle-wise\\
    \hline\hline
    MCTS & 55.63 \% & 14.55 \%  \\
    Greedy $P$ & 42.0 \% & 6.0 \% \\
    Greedy $V$ & 44.0 \% & 8.6 \% \\
    \hline
\end{tabular}
\end{center}
  \caption{Reassembly scores: comparison with and without MCTS.}
  \label{tab:azr:greedy}
\end{table}

\subsection{Alphazzle performances}

We propose several changes to our algorithm to improve Alphazzle's performances.

\subsubsection{Order of the patches}

We expect the order of the patches fed to MCTS to have a consequent impact on the reassembly score, and run an experiment to validate our hypothesis. We find out that solving $10$ times the same puzzle while reordering the patches and letting $V$ select its favorite leads to a gain of $14$\% on patches reassembly score and $30$\% on puzzle reassembly score. To compare, if we always select the worst reassembly on $10$ attempts, we obtain $33.48$\% and $3.43$\%. However, the possibility of changing the order of the patches goes hand in hand with an increase in computing time, limiting the number of reassemblies we made.


\subsubsection{Action choice from MCTS output}

MCTS returns the policy $\pi_{MCTS}(a_t|s_t)$. To select the action to perform, we use $N(a|s_t)$.  If we replace it by $Q(a|s_t)$, we observe a slight improvement from $55.63\%$ to $58.11\%$ for patches accuracy, and from $14.55\%$ to $16.75\%$ for puzzles accuracy. We observe that on $3\times3$ puzzles, $Q(a|s_t)$ is better than $N(a|s_t)$. Note that the puzzles are not identical from one generation to another, which explains why $Q(a|s_t)$  may have a lower puzzle-wise score. On bigger puzzles, it is better to prefer $N(a|s_t)$.


\subsubsection{Impact of fine-tuning}
We study the impact of fine-tuning P and V one states explore by MCTS. The complete results are available in appendix.
It appears that fine-tuning leads to a consistent improvement of results, albeit at a very slow rate.
Thanks to the fine-tuning, we have been able to double the number of well-solved puzzles.

\subsection{Reassembly results}
\label{sec:azr:reass}


\subsubsection{Qualitative analysis}

\begin{figure}
    \centering
    \subfloat[\label{fig:azr:0}]{\includegraphics[width=0.30\textwidth]{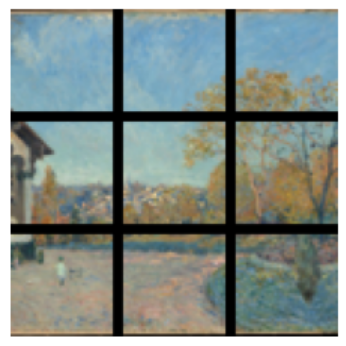}}\hfill
    \subfloat[\label{fig:azr:1}]{\includegraphics[width=0.30\textwidth]{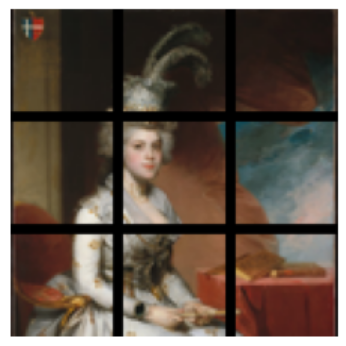}}\hfill
    \subfloat[\label{fig:azr:2}]{\includegraphics[width=0.30\textwidth]{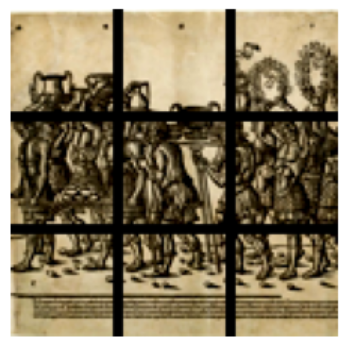}}
    
    \subfloat{\includegraphics[width=0.30\textwidth]{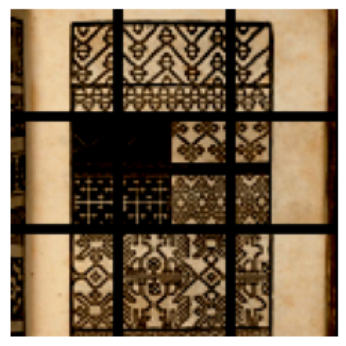}}\hfill
    \subfloat{\includegraphics[width=0.30\textwidth]{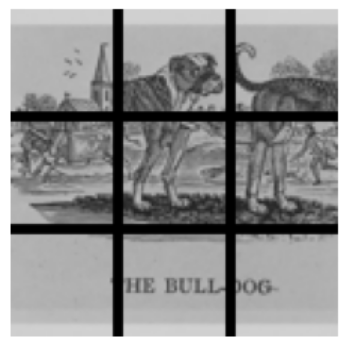}}\hfill
    \subfloat{\includegraphics[width=0.30\textwidth]{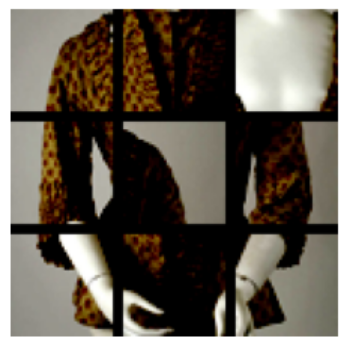}}\hfill
    
    \subfloat[\label{fig:azr:r1}]{\includegraphics[width=0.30\textwidth]{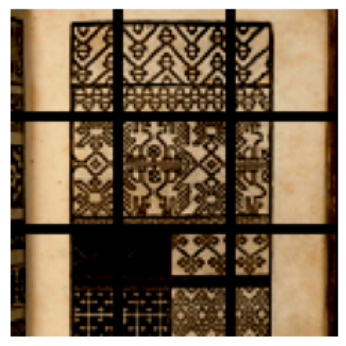}}\hfill
    \subfloat[\label{fig:azr:r5}]{\includegraphics[width=0.30\textwidth]{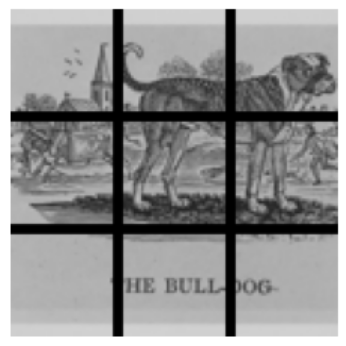}}\hfill
    \subfloat[\label{fig:azr:r4}]{\includegraphics[width=0.30\textwidth]{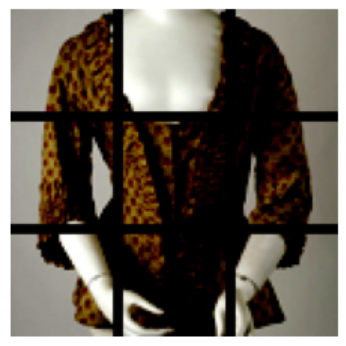}}\hfill
    
    \caption{Some reassemblies, with their solutions.}
    \label{fig:azr:results}
\end{figure}

Figure \ref{fig:azr:results} shows typical examples of correct and incorrect reassemblies. At first glance, we observe that most of the $3\times3$ reassemblies we made feature many well-placed patches. Indeed, on such a setup, puzzles with at most 2 mistakes represent more than $30$\% of the results.

Puzzles \subref{fig:azr:0}, \subref{fig:azr:1}, and \subref{fig:azr:2} are correct reassemblies. Puzzle \subref{fig:azr:r1} is visually correct, but the rows are inverted.
Puzzle \subref{fig:azr:r5} features an error of $V$: the head and the tail are swapped, despite being placed during the last steps.
Puzzle \subref{fig:azr:r4} illustrates how a misplaced first patch (the cleavage) alters the rest of the puzzle.

\subsubsection{Quantitative analysis}

We combine the standard experiment with fine-tuning, $Q(a|s_t)$ and 10 attempts. We reach 75.12\% (patch-wise score), 77.54\% (neighbor-wise score) and 51.49\% (puzzle-wise score). We also analyzed the errors distribution, depending on how many patches are well-placed. Most of the time, two or four patches are swapped.

Table \ref{tab:azr:comparison} displays the puzzle-wise score for different reassembly algorithms. The number of permutations refers to Noroozi and Favaro maximum number of classes. In their algorithm, the input is a series of patches and the output is one of the available permutations. We did not implement a way to limit the available permutations; therefore, we do not compare to the lowest numbers of available permutations.

\begin{table}
\begin{center}
  \begin{tabular}{|l|c|c|c|}
    \hline
    & \multicolumn{3}{c|}{\# of available permutations} \\
    Algorithm & $10^2$ & $10^3$ & $9!\simeq 10^6$ \\
    \hline\hline
    Noroozi and Favaro \cite{noroozi2016unsupervised} & 69.3 \% & 51.6 \% & overflow \\
    Paumard et al. \cite{paumard2020deepzzle} & 81.7 \% & 64.8 \% & 39.2 \%\\
    Ours & n/a & n/a & 51.5 \% \\
    \hline
  \end{tabular}
\end{center}
  \caption{Puzzle-wise scores.}
  \label{tab:azr:comparison}
\end{table}

We outperform \cite{paumard2020deepzzle} by $12.3\%$ on the puzzle accuracy. In terms of computational cost, our method greatly outperforms \cite{carlucci2019domain, noroozi2016unsupervised, paumard2020deepzzle}. Not only do we address all possible $n!$ permutation, but our strategy can tackle reassemblies to $4\times 4$ and $5 \times 5$ puzzles, whereas state of the art would require too much time to assemble such puzzles.

\section{Conclusion}

In this paper, we present a new algorithm, and we make three significant contributions.
First, we introduce a new method to solve bigger puzzles with deep learning and Monte-Carlo Tree Search. We propose an algorithm inspired by AlphaZero's MCTS, that extracts visual features and has no access to the reward. 
Second, our method outperforms state of the art in terms of computational cost and accuracy of the reassembly.
Third, we run an in-depth analysis of the parameters of our algorithm. 
Among other things, we demonstrate that predicting the reward is much easier than predicting the best action. We also show that controlling the order of the patches and fine-tuning the networks on failed reassemblies greatly improve the results.


\bibliography{mybibfile}

\appendix
\section*{Appendix}

\section{Pre-training performances}

Table \ref{tab1}. shows the impact of settings on neural networks. First of all, we note it is easier for $V$ to scout mistakes in the reassembly than for $P$ to predict the action.

\paragraph{Fragment per side} Increasing the number of fragments in a puzzle does not profoundly impact the performance of $V$ but leads to a drop in performance for the network $P$, which is not compensated when training with already placed fragments. In Table \ref{tab1}, 6×6 puzzles reach a validation accuracy of 3.43\% for $P$ and of 64.47\% for $V$.

\paragraph{Space size} Increasing the size of the space between the fragments makes the puzzles slightly more complicated, by a few percent for $P$ and $V$. It means our neural networks successfully learn to solve puzzles without relying on continuities.

\paragraph{Fragment size} We ran very few experiments on the fragment size. Note that on 96×96 fragments, the space size is half the fragment size. Therefore, we should compare the results to the 40×40 fragments spaced by 20 pixels. According to Table \ref{tab1}, bigger fragments lead to -5\% accuracy for $P$ and $V$, probably because our architecture is not well suited for larger fragments, or at least not well tuned for them.

\paragraph{Placed fragments} The placed fragments, or “hints”, are the minimal number of placed fragments in the reassembly when training or evaluating the networks. For $P$, they are the well-placed fragments; for $V$, correct reassemblies alternates with reassemblies that can lead to a wrong reassembly. The impact of hints on results allows studying how our neural networks behave on easier tasks. For example, when we give 8 hints to a neural network that solves the 4×4 puzzle, we theoretically approximate the 3×3 puzzle difficulty, but the performance is lower on $P$.

We also consider a central hint to compare with Deepzzle. When there are 8 hints, it means that there is only one fragment to place in the case of 3×3 puzzles. The results show that the neural network $P$ successfully learned to place the fragments in empty positions. However, it did not reach 100\% because some background fragments are as black as empty spaces. Inspired by AlphaGo’s data structure, we proposed to append a 4th channel to the images that allow differentiating empty spaces and black fragments. We did not see any visible improvement of the validation accuracy, but the computing time has become slightly longer.

We ran more experiments to analyze the impact on hints. We present the results in Table \ref{tab2}. In this Table, contrarily to the previous one, the validation hints does not indicate the minimal number of hints, but their exact number.

Giving hints makes the neural networks perform better in solving easy puzzles (with as many or more hints). However, the validation accuracy we obtain dropped significantly on puzzles with fewer hints than during the training: we obtain 11.34\% $\simeq$ 1/9 on empty reassembly if we train $P$ to place only the last fragment. Note that the standard $P$ network trained with no hint can correctly predict the position of the first fragment 46.97\% of the time.

If we study the impact of hints on specific partial reassembly, for every amount of fragments placed, we see that performance on easy puzzles (i.e., when most fragments are placed) are equivalent for various minimal numbers of training hints. It means that using specialized neural networks\footnote{As suggested in earlier versions of AlphaGo, with their different $P$ networks for opening, middle-game, and endgame.} for the last steps brings no substantial gain. A neural network which has not been trained on difficult partial reassemblies performs worse than a neural network trained on all type of partial reassemblies. However, it will still be better than random on puzzles slightly more difficult than those on which it learned. Last but not least, $V$ displays bad accuracies on the most challenging reassemblies. It is because it learns to express its uncertainty about the puzzles: it makes soft guesses over the class.

\begin{table*}
\begin{center}
  \begin{tabular}{|l|l|l|l|l|r|r|}
    \hline
    \multicolumn{5}{|c|}{Configuration} & \multicolumn{2}{|c|}{Validation accuracy}            \\
    \hline
    \shortstack{Patch \\ size (px)} & \shortstack{Patch \\ per side} & \shortstack{Space size \\ (px)} & \shortstack{Hints \\ training} & \shortstack{Hints \\ validation} & P (\%) & V (\%) \\
    \hline\hline
    40 & 3 & 4 & 0 & 0 & 69.91 & 88.46 \\
    40 & 3 & 4 & 0 & 4 & 79.43 & 95.51 \\
    40 & 3 & 4 & 0 & 8 & 99.29 & 97.93 \\
    40 & 3 & 4 & 1 (central) & 1 (central) & 73.99 & 92.34 \\
    40 & 3 & 4 & 2 & 2 & 71.82 & 93.35 \\
    40 & 3 & 4 & 4 & 0 & 65.52 & 86.36 \\
    40 & 3 & 4 & 4 & 4 & 79.98 & 95.77 \\
    40 & 3 & 4 & 4 & 8 & 99.74 & 95.48 \\
    40 & 3 & 4 & 6 & 6 & 87.50 & 97.53 \\
    40 & 3 & 4 & 8 & 0 & 29.35 & 70.96 \\
    40 & 3 & 4 & 8 & 4 & 42.59 & 82.11 \\
    40 & 3 & 4 & 8 & 8 & 99.49 & 88.45 \\
    40 & 3 & 10 & 0 & 0 & 67.39 & 87.15 \\
    40 & 3 & 20 & 0 & 0 & 61.34 & 85.79 \\
    \hline
    40 & 4 & 4 & 0 & 0 & 37.65 & 92.64 \\
    40 & 4 & 4 & 4 & 4 & 48.54 & 98.24 \\
    40 & 4 & 4 & 8 & 8 & 52.02 & 99.09 \\
    40 & 4 & 4 & 12 & 12 & 72.53 & 99.04 \\
    40 & 4 & 20 & 0 & 0 & 39.67 & 90.47 \\
    40 & 5 & 4 & 0 & 0 & 19.15 & 94.25 \\
    40 & 5 & 4 & 10 & 10 & 23.79 & 99.50 \\
    40 & 6 & 4 & 0 & 0 & 3.43 & 64.47 \\
    \hline
    96 & 3 & 48 & 0 & 0 & 56.55 & 80.54 \\
    96 & 3 & 48 & 1 (central) & 1 (central) & 60.89 & 73.08 \\
    96 & 3 & 48 & 2 & 2 & 67.24 & 88.00 \\
    96 & 3 & 48 & 4 & 4 & 75.30 & 90.02 \\
    \hline
    40 & 2 & 0 & 2 & 2 & 86.84 & 82.66 \\
    24 & 3 & 4 & 5 & 5 & 79.44 & 95.67 \\
    \hline
  \end{tabular}
\end{center}
\caption{Validation accuracy for $P$ and $V$ on various configurations}
\label{tab1}
\end{table*}

\begin{table*}
\begin{center}
  \begin{tabular}{|l|l|l|l|l|r|r|}
    \hline
    \multicolumn{5}{|c|}{Configuration} & \multicolumn{2}{|c|}{Validation accuracy}            \\
    \hline
    \shortstack{Patch \\ size (px)} & \shortstack{Patch \\ per side} & \shortstack{Space size \\ (px)} & \shortstack{Hints \\ training} & \shortstack{Hints \\ validation} & P (\%) & V (\%) \\
    \hline\hline
    40 & 3 & 4 & 0 & 0 patch & 46.97 & 50.00 \\
    40 & 3 & 4 & 0 & 1 patch & 51.96 & 69.35 \\
    40 & 3 & 4 & 0 & 2 patches & 56.25 & 85.78 \\
    40 & 3 & 4 & 0 & 3 patches & 57.71 & 88.91 \\
    40 & 3 & 4 & 0 & 4 patches & 63.45 & 90.57 \\
    40 & 3 & 4 & 0 & 5 patches & 69.10 & 92.64 \\
    40 & 3 & 4 & 0 & 6 patches & 77.37 & 96.02 \\
    40 & 3 & 4 & 0 & 7 patches & 85.33 & 96.97 \\
    40 & 3 & 4 & 0 & 8 patches & 99.49 & 99.29 \\
    40 & 3 & 4 & 0 & 9 patches & --- & 96.77 \\
    40 & 3 & 4 & 4 & 0 patch & 39.61 & 50.00 \\
    40 & 3 & 4 & 4 & 1 patch & 46.98 & 70.97 \\
    40 & 3 & 4 & 4 & 2 patches & 52.77 & 86.49 \\
    40 & 3 & 4 & 4 & 3 patches & 58.62 & 88.71 \\
    40 & 3 & 4 & 4 & 4 patches & 63.61 & 91.33 \\
    40 & 3 & 4 & 4 & 5 patches & 70.46 & 93.19 \\
    40 & 3 & 4 & 4 & 6 patches & 76.46 & 96.37 \\
    40 & 3 & 4 & 4 & 7 patches & 84.53 & 97.88 \\
    40 & 3 & 4 & 4 & 8 patches & 99.65 & 98.94 \\
    40 & 3 & 4 & 4 & 9 patches & --- & 98.23 \\
    40 & 3 & 4 & 8 & 0 patch & 11.34 & 50.00 \\
    40 & 3 & 4 & 8 & 1 patch & 11.99 & 50.00 \\
    40 & 3 & 4 & 8 & 2 patches & 11.74 & 52.47 \\
    40 & 3 & 4 & 8 & 3 patches & 16.78 & 57.86 \\
    40 & 3 & 4 & 8 & 4 patches & 17.79 & 66.73 \\
    40 & 3 & 4 & 8 & 5 patches & 23.18 & 75.30 \\
    40 & 3 & 4 & 8 & 6 patches & 30.79 & 79.99 \\
    40 & 3 & 4 & 8 & 7 patches & 49.54 & 82.66 \\
    40 & 3 & 4 & 8 & 8 patches & 99.39 & 86.59 \\
    40 & 3 & 4 & 8 & 9 patches & --- & 91.23 \\
    \hline
  \end{tabular}
\end{center}
  \caption{Validation accuracy for all type of partial reassemblies}
  \label{tab2}
\end{table*}

\section{MCTS meta-parameter optimization}
We analyze the number of visits $N_{visits}$ to run before selecting the action, and the trade-off $C$ between exploration and exploitation. We want to select the best configuration for the reassembly task.

Table \ref{tab:grid-search} presents some results on meta-parameters. We obtain the best score with a high number of simulations and $C=1$. Running many simulations increases the computation time drastically, although results are always better with more simulations. Note that high exploration (when $C>1$) also has a non-negligible computational cost as more new states have to be analyzed by the neural networks. For that reason, we use $10^3$ simulations and $C=1$ in the following experiments. 

\begin{table*}
\begin{center}
  \begin{tabular}{|r|r|r|r|r|r|r|r|r|r|}
    \hline
    $N_{visits}$ & 10 & $10^2$ & $10^3$ & $10^3$& $10^3$ & $10^3$ & $10^4$ & $10^5$ & $10^6$ \\
    $C$ & 1 & 1 & 0.01 & 0.1 & 1 & 10 & 1 & 1 & 1 \\
    \hline\hline
    Fragment accuracy (\%) & 49.0 & 54.5 & 48.8 & 55.0 & 55.6 & 52.2 & 57.9 & 58.0 & 65.6 \\
    Puzzle accuracy (\%) & 12 & 15 & 10 & 14 & 15 & 13 & 17 & 22 & 30 \\
    Solving time (s/puzzle) & 1 & 4 & 8 & 9 & 16 & 25 & 84 & 685 & 7200 \\
    \hline
  \end{tabular}
  \end{center}
  \caption{Reassembly scores --- Comparison of MCTS meta-parameters.}
  \label{tab:grid-search}
\end{table*}

\section{Deactivation of $P$ and $V$}

We already analyzed the results of Table \ref{tab:reass-pvv} in the article.

\begin{table*}
\begin{center}
  \begin{tabular}{|l|l|l|l|l|l|l|l|l|}
    \hline
    P & $\checkmark$ & $\checkmark$ & $\checkmark$ & $\checkmark$ & & & & \\ 
    V (endgame) & $\checkmark$ & $\checkmark$ & & & $\checkmark$ & $\checkmark$ & & \\ 
    V (during game) & $\checkmark$ & & $\checkmark$ & & $\checkmark$ & & $\checkmark$ & \\ 
    \hline\hline
    Fragment accuracy (\%) & 55.63 & 53.64 & 49.00 & 45.01 & 54.17 & 14.47 & 51.47 & 11.23 \\ 
    Puzzle accuracy (\%) & 14.55 & 14.65 & 11.10 & 9.20 & 12.25 & 0.00 & 10.50 & 0.00 \\ 
    \hline
  \end{tabular}
  \end{center}
  \caption{Reassembly scores --- Comparison of MCTS with and without $P$ or $V$.}
  \label{tab:reass-pvv}
\end{table*}

\section{Order of the fragments}

We expect the order of the fragments fed to MCTS to have a consequent impact on the reassembly score, and run an experiment to validate our hypothesis.

Table \ref{tab:azr:nb-tirages} details the impact of making several attempts to solve a puzzle, with different input reorderings. We find out that solving \oldstylenums{10} times the same puzzle while reordering the fragments and letting $V$ selecting its favorite leads to a gain of \oldstylenums{14}\% on fragments reassembly score and \oldstylenums{30}\% on puzzle reassembly score. To compare, if we always select the worst reassembly on \oldstylenums{10} attempts, we obtain \oldstylenums{33.48}\% and \oldstylenums{3.43}\%, which was lucky compared to the worst attempt we made on the \oldstylenums{5} attempts test.

\begin{table*}
\begin{center}
  \begin{tabular}{|l|r|r|r|r|r|}
    \hline
     &
    \multicolumn{2}{c}{Best attempt} &
    \multicolumn{2}{c}{Worst attempt} & \\
    \hline
    \shortstack{Number \\ of attempt} & 
    \shortstack{Fragment-\\wise (\%)} & \shortstack{Puzzle-\\wise (\%)} &
    \shortstack{Fragment-\\wise (\%)} & \shortstack{Puzzle-\\wise (\%)} & \shortstack{Reassemblies\\done in 24h}\\
    \hline\hline
    1 & 55.63 & 14.55 & - & - & 2000 \\
    5 & 64.49 & 42.33 & 24.20 & 0.90 & 2000 \\
    10 & 68.98 & 44.59 & 33.48 & 3.43 & 466 \\
    20 & 70.62 & 41.69 & 36.04 & 1.80 & 222 \\
    \hline
  \end{tabular}
  \end{center}
  \caption{Reassembly scores --- Impact of the order of fragments.}
  \label{tab:azr:nb-tirages}
\end{table*}

However, the possibility of changing the order of the fragments goes hand in hand with an increase in computing time, limiting the number of reassemblies we made.

When we use multiple fragments order during the inference, we select the best solution according to $V$ rather than the true best solution. Therefore, the final scores may be lower than those displayed in Table \ref{tab:azr:nb-tirages}.

\section{Action choice from MCTS output}

We already analyzed the results of Table \ref{tab:azr:qsa} in the article.

\begin{table*}
\begin{center}
  \begin{tabular}{|l|l|l|r|r|r|r|}
    \hline
    & & &
    \multicolumn{2}{c}{$N(a|s_t)$} &
    \multicolumn{2}{c}{$Q(a|s_t)$} \\
    \hline
    \shortstack{Fragment \\ per side} & 
    \shortstack{Hints \\ reassembly} &
    \shortstack{Number \\ of attempts} &
    \shortstack{Fragment-\\wise (\%)} & \shortstack{Puzzle-\\wise (\%)} & \shortstack{Fragment-\\wise (\%)} & \shortstack{Puzzle-\\wise (\%)} \\
    \hline\hline
    3 & 0 & 1 & 55.63 & 14.55 & 58.11 & 16.75 \\
    3 & 0 & 10 & 64.49 & 42.33 & 72.28 & 39.56 \\
    3 & 1 (central) & 10 & 80.66 & 45.95 & 82.47 & 51.50 \\
    \hline
    4 & 0 & 10 & 35.32 & 0.74 & 32.49 & 0.76  \\
    5 & 0 & 10 & 15.92 & 0.00 & 14.04 & 0.00  \\
    \hline
  \end{tabular}
  \end{center}
  \caption[Reassembly --- Comparison between $Q(a|s_t)$ and $N(a|s_t)$]{Reassembly scores --- Impact of $Q(a|s_t)$.}
  \label{tab:azr:qsa}
\end{table*}

\section{Impact of fine-tuning}

Table \ref{tab:azr:reass-finetuning-a} shows the fine-tuning results on various epochs sizes, without using other optimization such as the order of fragments. Note that we make 100 simulations rather than 1000, leading MCTS to make more mistakes, which should provide more relevant training samples. After a few epochs, we significantly improve the puzzle reassembly scores by almost $9\%$. 

\begin{table*}
\begin{center}
  \begin{tabular}{|l|r|r|r|r|r|r|r|r|}
    \hline
    Fine-tuning epoch & 0 & 1 & 3 & 5 & 7 & 10 & 15 & 20\\
    \hline\hline
    Puzzle accuracy (\%)& 14.55 & 17.35 & 18.25 & 20.05 & 20.40 & 20.00 & 20.30 & 23.10\\
    \hline
  \end{tabular}
\end{center}
  \caption{Reassembly scores --- Detail on fine-tuning accuracy for $3\times3$ puzzles with 500 puzzles generated per iteration.}
  \label{tab:azr:reass-finetuning-a}
\end{table*}

\end{document}